\def\year{2022}\relax
\documentclass[letterpaper]{article} 
\usepackage{aaai22}  
\usepackage{times}  
\usepackage{helvet}  
\usepackage{courier}  
\usepackage[hyphens]{url}  
\usepackage{graphicx} 
\usepackage{natbib}  
\usepackage{caption} 
\DeclareCaptionStyle{ruled}{labelfont=normalfont,labelsep=colon,strut=off} 
\usepackage{amsmath}
\usepackage[ruled,vlined]{algorithm2e}
\usepackage{multirow}
\usepackage{tabularx}
\usepackage{array}
\usepackage{tcolorbox}
\usepackage[toc,page]{appendix}
\usepackage{tabu}
\usepackage{booktabs}

\DeclareMathOperator*{\argmax}{argmax}

\title{A Bayesian Approach for Medical Inquiry and Disease Inference\\ in Automated Differential Diagnosis}
\author{
    Hong Guan, Chitta Baral
    \\
}
\affiliations{
    Arizona State University
}

\begin{document}
\maketitle
\begin{abstract}
We propose a Bayesian approach for both medical inquiry and disease inference, the two major phases in differential diagnosis. Unlike previous work that simulates data from given probabilities and uses ML algorithms on them, we directly use the Quick Medical Reference (QMR) belief network, and apply Bayesian inference in the inference phase and Bayesian experimental design in the inquiry phase. Moreover, we improve the inquiry phase by extending the Bayesian experimental design framework from one-step search to multi-step search. Our approach has some practical advantages as it is interpretable, free of costly training, and able to adapt to new changes without any additional effort. Our experiments show that our approach achieves new state-of-the-art results on two simulated datasets, SymCAT and HPO, and competitive results on two diagnosis dialogue datasets, Muzhi and Dxy.
\end{abstract}

\section{Introduction}
In healthcare, differential diagnosis is a process wherein a doctor analyzes patient information such as symptoms, signs and results of medical tests to identify a particular disease and differentiate it from others that may have similar clinical findings. Automation of this process, 
referred to as ``automated differential diagnosis'', is a crucial component of the clinical decision support system (CDSS) \cite{ledley1959reasoning,10.1136/jamia.1994.95236141,sutton2020overview}, that supports clinicians in making diagnostic decisions. In healthcare practice, differential diagnosis involves several aspects, such as: decision trees and utility functions  to decide what test to take \cite{biomedical_informatics}, and analysis of medical imaging and radiology \cite{ai_imaging} as part of tests. Despite the complexity of differential diagnosis, in this paper we abstract differential diagnosis to a simplified problem, which can be easily extended to more general cases. 

More specifically, we consider the following scenario. A patient goes to see a doctor and informs the doctor his/her symptoms, and the doctor may want to ask the patient some questions or perform some clinical tests to gather more information. When the collected information is enough, the doctor makes a final diagnosis. We model the doctor’s questions as inquiry steps and the final diagnosis as the final inference step. From a computer system’s perspective, the system is first given some initial findings, and then continues to inquire about some new findings, and finally predicts a disease as a final diagnosis. In this framework, health history, physical exams, radiology results can also be included as findings. 

Machine learning approaches such as decision tree, Naive Bayes, neural networks, Bayesian inference have been used to predict a disease given a set of known findings \cite{kononenko2001machine,Jaakkola_1999}. However, such systems can not actively acquire new information or suggest new clinical tests. Recent work combines inquiry and inference into a reinforcement learning (RL) framework \cite{tang2016inquire,kao2018context,refuel}. Another class of methods is based on partial variational autoencoder (partial VAE) \cite{ma2019eddi,he2020fit}, which uses one neural network to perform Bayesian experimental design \cite{10.1214/ss/1177009939} for inquiry and another to perform disease classification, and jointly trains these two networks. However, there are some limitations in these two methods. RL-based methods are computational intensive and have poor control of the inquiry and inference process. Partial VAE-based methods usually need many inquiry steps to achieve high accuracy. And both of these two methods need to re-train the neural networks when the data changes, e.g. a new disease or finding is added to the system. 

In this work, we propose a Bayesian approach that is free of the above limitations. In particular, we use the QMR belief network \cite{pmid3537611} as the graphical model, and develop an inference engine that can make inference given arbitrary known findings, and then propose an inquiry strategy based on Bayesian experimental design, which maximizes the information gain predicted by the inference engine. 
{\em Our contributions} include applying Bayesian experimental design to the QMR belief network and extending Bayesian experimental design from one-step search to multi-step search. Moreover, in achieving new state-of-the-art performance on two simulated datasets and competitive results on two dialogue datasets, our methods have the following advantages: they are highly scalable and do not require costly training; the active inquiry and inference phases can easily adapt to manual interference and new changes in database or knowledge base; and the results can be interpreted by inspecting the graphical structure of the QMR belief network.
\section{Related works}
There are some early works that build the association between diseases and findings and use it for automated diagnosis. The INTERNIST-1/QMR project \cite{pmid3537611,pmid3544509} builds an expert database QMR (Quick Medical Reference) describing 570 diseases in internal medicine. \cite{pmid1762578,pmid1743005} reformulate the QMR database as a bipartite belief network, which is called QMR-DT (decision theoretic). The inference problem of QMR-DT has been studied in \citet{Jaakkola_1999}, \citet{ng1999approximate}, \citet{feili2003} and \citet{yu_2007}. However, these methods can not actively acquire new information, i.e. they do not consider the inquiry problem.

\citet{tang2016inquire} formulates the inquiry and inference problem in a reinforcement learning framework. \citet{kao2018context} adds patient information such as age, gender, and season as contextual information into their model. \citet{refuel} uses reward shaping and feature rebuilding to encourage better and faster learning. In these methods, the state of the environment consists of all the states of the findings, either present or absent or unknown. Each finding and disease is modeled as one action. Choosing a finding action means inquiring about that finding and changing the state according to the result of the inquiry, either present or absent. And choosing a disease action means making a final diagnosis and terminating the whole process. Although it is theoretically plausible, the action space of such methods is very big, which makes the learning difficult and slow. Moreover, they will fail in some cases where many inquiry steps are taken without achieving a diagnosis when the system must stop at an allowed maximum step. This is mainly due to the absence of a separate inference engine. Instead, our method combines all diseases as one diagnosis step and uses a separate inference engine to make the final diagnosis.

EDDI \cite{ma2019eddi} is based on Bayesian experimental design \cite{10.1214/ss/1177009939} and partial VAE, where they train a neural network to find the latent variables of the partially observed data. FIT \cite{he2020fit} improves EDDI by improving the neural network architecture and using a separate neural network for disease classification, a faster sample strategy, and other speed-up tricks. These two approaches can achieve similar or better accuracy compared to reinforcement learning based methods, with some compromise on the number of inquiry steps for some cases. However, they extract patterns from the data but do not take advantage of the graphical model that generates the data. 

In this work, we have an inference model and an inquiry model, and the inference model is used in every inquiry step. More specifically, inference is achieved by Bayesian inference on the QMR belief network, similar to the work in \citet{Jaakkola_1999}. During the inquiry phase, unlike EDDI \cite{ma2019eddi} and FIT \cite{he2020fit}, in which Bayesian experimental design is applied to the latent space of the data, we directly apply Bayesian experimental design to the inference result.
\section{Background}
In this section, we first describe the QMR belief network \cite{pmid1762578}, a model to represent the association between diseases and findings. Then we review the derivation of Bayesian inference for a QMR belief network as in \cite{Jaakkola_1999}. It can be considered as the last step of the whole diagnosis process, by making a final diagnosis given all the available information, i.e. absence or presence of some findings. To help make a better final diagnosis, one may want to acquire more information about findings given the partial observation. Lastly, Bayesian experimental design \cite{10.1214/ss/1177009939}, a framework we use to guide the inquiry strategy, will be reviewed.

\subsection{QMR belief network}
\begin{figure}[htp]
    \centering
    \includegraphics[width=8cm]{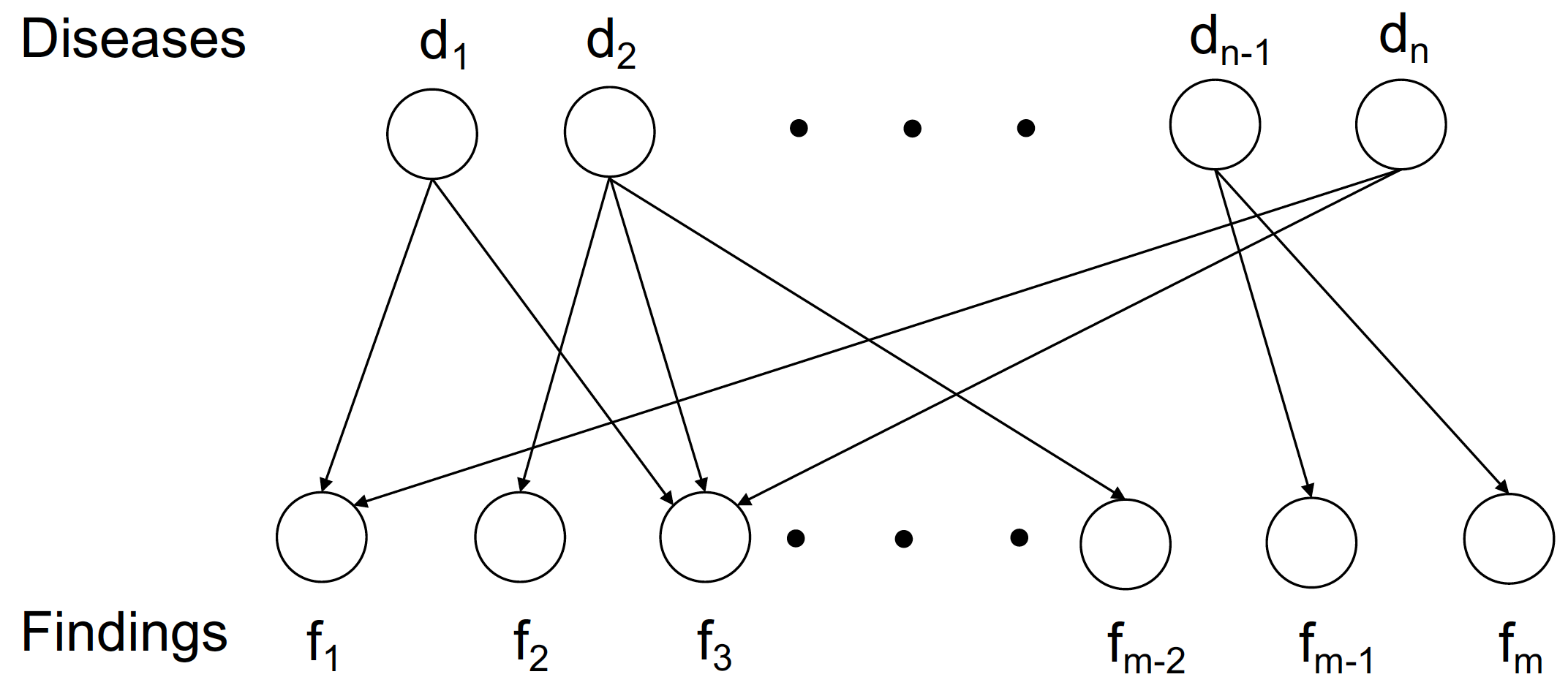}
    \caption{The QMR belief network}
    \label{fig:qmr}
\end{figure}

Figure \ref{fig:qmr} shows a QMR belief network \cite{pmid1762578} with $n$ diseases and $m$ findings, which is a two-level graph describing the association between diseases and findings (including symptoms, signs, and other useful information for diagnosis). Here, we describe the QMR belief network following the convention in \cite{Jaakkola_1999}. Each node takes a value 0 or 1, representing absence and presence respectively. Each disease has a positive number (e.g. $P(d_j=1)$ for the $j$th disease) representing the prior belief of the marginal probability of a disease. In this work, we consider one and only one disease per case, thus we assume the sum of marginal probabilities of all diseases to be one. Each edge has a positive number representing the probability that if the pointing disease presents, it can alone cause the presence of the pointed finding. We denote the number linking the $i$th finding and the $j$th disease as $P(f_i=1|d_j=1)$ and use  $P(f_i|d_j)\equiv P(f_i\mathop{=}1|d_j\mathop{=}1)$ for the sake of brevity. Since multiple diseases can cause a finding, the interaction in this conjunction is modeled by a nosiy-OR gate \cite{pearl1988}. In other words, a finding is negative if and only if its parent diseases are all negative. 

\subsection{Disease Inference, the general case}
In differential diagnosis, the quantity one cares about is the probabilities of each disease given some findings, i.e. $P(d_j=1|f^+,f^-)$ for $j=0,1,2,\dots,n$, where $f^+$ denotes a set of positive findings and $f^-$ denotes a set of negative findings. If the patient has the symptom, the finding is positive, if the patient does not have the symptom the finding is negative. Since the disease nodes are the parents of finding nodes, one can apply Bayes Theorem to $P(d_j=1|f^+,f^-)$,
\begin{align}
    P(d_j=1|f^+,f^-) &= \frac{P(f^+,f^-,d_j=1)}{P(f^+,f^-)} \\
    &\propto \displaystyle\sum_{d\backslash d_j}P(f^+,f^-|d)P(d)
    \label{eq:multi_inference}
\end{align}
where $d$ is a vector of all diseases, taking either value 0 or 1. The notation $d \backslash d_j$ is a shorthand for summarizing all diseases except $d_j$. Note that the term $P(f^+,f^-|d)$ can be factorized because the findings are conditionally independent given all diseases. Thus,
\begin{equation}
     P(f^+,f^-|d)P(d) =  \displaystyle\prod_{i}P(f_i^+|d)\displaystyle\prod_{k}P(f_k^-|d)P(d) \label{eq:multi-expansion}
\end{equation}
The above derivation is adopted from \cite{Jaakkola_1999}. The cost of computing the summation is exponential to the number of diseases. In the case of n diseases, the complexity is around $2^{n-1}$ per disease. 

\subsection{Bayesian Experimental Design} \label{bed}
Bayesian experimental design \cite{10.1214/ss/1177009939} is a general probability-theoretical approach to help direct experimental design. EDDI \cite{ma2019eddi} and FIT \cite{he2020fit} simulate the data using the QMR belief network and apply this approach to the latent space. However, in this work, we directly compute the probability distribution of all the diseases using Bayesian inference as previously mentioned, and apply Bayesian experimental design to the diseases.

Here is our problem definition in the framework of Bayesian experimental design. Let $\theta$ denotes the ``target'', a patient's true disease; let $\xi$ denotes the ``experiment'', checking if a finding exists; and let $y$ denotes the ``outcome'', whether a finding exists. All of the above variables take binary values 0 and 1. The objective is to find $\xi_{obj}\mathop{=}\argmax_{\xi}U(\xi)$, i.e. the experiment (question) that maximize the utility function $U(\xi)$ given by 
\begin{equation}
    U(\xi)=\int p(y|\xi)U(y,\xi)\mathrm{d}y \label{eq:util}
\end{equation}

Intuitively, equation \ref{eq:util} computes the utility of an experiment $\xi$, one would want to add up the utilities of all possible outcomes $y$. Two common choices of the utility function $U(y,\xi)$ are Kullback–Leibler (KL) divergence and prior-posterior gain in Shannon information (IG). Intuitively, these two functions measure how much an experiment (a question) changes one's belief of the target (true disease). When taking KL divergence as the utility function, it can be written as the follows,
\begin{align}
    U(y,\xi)&= D_{\mathrm{KL}}\big(p(\theta|y,\xi)\|p(\theta)\big) \nonumber \\
    &= \log\big(p(\theta|y,\xi)\big)p(\theta|y,\xi) - \log\big(p(\theta|y,\xi)\big)p(\theta)
    \label{eq:kl}
\end{align}
and when taking IG as the utility function, it becomes
\begin{align}
        U(y,\xi) &= IG\big(p(\theta|y,\xi),p(\theta)\big) \nonumber \\
        &= \log\big(p(\theta|y,\xi)\big)p(\theta|y,\xi) - \log\big(p(\theta)\big)p(\theta) \label{eq:ig}
\end{align}

These two utility functions are similar and output very close results in our experiments. For the sake of simplicity, we only use KL divergence in describing our approach.
\section{Proposed Method}

\subsection{Task Description}
In this work, we assume that the state of a finding is fully known by the patient but is partially observed by our system. The state of a finding is either present/positive or absent/negative. Initially, the patient reports some findings, either positive or negative. 
At each step of the inquiry process, our system is asked to output a finding to check if the finding is positive or negative. And we assume that the patient will correctly inform the system.
At the end of the inquiry process, given all the known findings, our system is asked to make a final prediction of what disease the patient may have, or more generally, to output the probabilities over all diseases. 

\subsection{Bayesian Disease Inference for One Disease per Case} \label{section:bed-one-disease}
The full inference problem for the QMR belief network has an exponential time complexity, which is impractical for many real-time healthcare applications. However, we show that by limiting the number of diseases per case to one, the time complexity is linear to the number of positive findings. This simplification does not reduce the usefulness of our method, because one disease per case is quite common in some healthcare practice.

Under the assumption of one disease per case, the $d_j=1$ in equation \ref{eq:multi_inference} implies $d_i=0$ for all $i \neq j$. Let $d^j$ denote the one hot vector with
\[ (d^j)_i =
  \begin{cases}
    1  & i = j ,\\
    0  & i \neq j
  \end{cases}
\]

Then equation \ref{eq:multi_inference} can be rewritten as
\begin{align}
P(d_j=1|f^+,f^-) =& \frac{P(f^+,f^-,d_j=1)}{P(f^+,f^-)} \nonumber \\
     = &\frac{P(f^+,f^-,d^j)}{P(f^+,f^-)}  \label{eq:bi-each-disease}
\end{align}
where
\begin{align}
P(f^+,f^-,d^j) =& \displaystyle\prod_{i}P(f_i^+|d^j)\cdot\displaystyle\prod_{k}P(f_k^-|d^j)\cdot P(d^j) \nonumber \\
     = &\displaystyle\prod_{i:f_i\in f^+}P(f_i|d^j)\cdot
     \displaystyle\prod_{k:f_k\in f^-}\big[1-P(f_k|d^j)\big]\cdot \nonumber \\ &P(d^j) \label{eq:bi-numerator} \\ 
P(f^+,f^-) =& \sum_{j}P(f^+,f^-,d^j) \label{eq:bi-denumerator}
\end{align}

For any $l$, $P(f_l|d^j)$ is the probability that if the $j$th disease present, can alone cause the $l$th finding to be positive. This probability can be found from the edges of the QMR belief network. Since we assume the diseases are mutually exclusive and the per case probabilities of the diseases add up to one, the resulting $P(f^+,f^-,d^j)$ looks like a Naive Bayes classifier. Compared to equation \ref{eq:multi_inference}, this equation avoids the summation over all diseases. The Bayesian inference algorithm is summarized as Algorithm \ref{alg:bi}. Note that the two \emph{for} loops are parallelizable, making it scalable.

\SetKwInput{kwInit}{Init}
\SetKwInput{kwSetup}{Setup}
\SetKwInput{kwInput}{Input}
\SetKwInput{kwOutput}{Output}

\begin{algorithm}
\SetAlgoLined
 \caption{Bayesian inference algorithm for differential diagnosis}
 \label{alg:bi}
\kwInput{Positive findings $f^+$ and negative findings $f^-$.}
 \For{$j$ in $0..n$}{
    Compute $P(f^+,f^-,d^j)$  using equation \ref{eq:bi-numerator}\;
 }
 Compute $P(f^+,f^-)$ using equation \ref{eq:bi-denumerator}\;
 Initialize an empty list $L$ to store the probability for each disease\;
 \For{$j$ in $0..n$}{
    Compute $P(d_j=1|f^+,f^-)$ using equation \ref{eq:bi-each-disease} and store it to the list $L$\; 
 }
 \kwOutput{Pick top k values from the list $L$ and output the corresponding diseases $d_{\mathrm{diagnosis}}$}
\end{algorithm}

\subsection{Finding Inquiry using Bayesian Experimental Design}
Often patients do not provide full information of the findings, in part because they must wait until further tests are taken in some situations. The differential diagnosis system should be able to actively inquire the patient to acquire more information to help the final diagnosis.  Intuitively, the Bayesian experimental design method can be interpreted as follows: the probability of every disease can be computed by Bayesian inference, which can be considered as the ``information'' one has; the objective of Bayesian experimental design is to maximize the probability of 	``information gain'' by inquiring about a particular finding.

Denote the newly found finding $f'$, then $p(y|\xi)$ is intuitively equivalent to $P(f'{=}y|f^+,f^-), y\in\{0,1\}$; $p(\theta|y,\xi)$ is intuitively equivalent to compute $P(d|f',f^+,f^-)$, that is, $P(d_{j}{=}\theta|f'{=}1,f^+,f^-), \theta\in\{0,1\}$ for all j; $p(\theta)$ is equivalent to compute $P(d|f^+,f^-)$, that is, $P(d_{j}{=}\theta|f^+,f^-), \theta\in\{0,1\}$ for all j. With these quantities and incorporating equation \ref{eq:util} and equation \ref{eq:kl}, we have 

\begin{align}
U(f')&=\displaystyle\sum_{y=0,1}P(f'{=}y|f^+,f^-)\sum_{j} \nonumber \\
& \;\;\;\;D_\mathrm{KL}(P(d_j|f^+,f^-,f'{=}y)||P(d_j|f^+,f^-)) \nonumber \\
&= \displaystyle\sum_{y=0,1}P(f'{=}y|f^+,f^-)\sum_{j}\sum_{\theta=0,1}\big[ \nonumber \\
& \;\;\log (P(d_j{=}\theta|f^+{,}f^-{,}f'{=}y))P(d_j{=}\theta|f^+{,}f^-{,}f'{=}y) \nonumber \\
& \;\;-\log (P(d_j{=}\theta|f^+,f^-,f'{=}y))P(d_j{=}\theta|f^+,f^-)\big] \label{eq:utility-kl}
\end{align}
where
\begin{align*}
P(f'=y|f^+,f^-) =& \frac{P(f'=y,f^+,f^-)}{P(f^+,f^-)} \\
    =& \frac{\sum_{j}P(f'=y,f^+,f^-,d^j)}{\sum_{j}P(f^+,f^-,d^j)}
\end{align*}. 

The objective at each step is to select the $f'$ that maximize $U(f')$. Algorithm \ref{alg:bed} summarizes the Bayesian experimental design algorithm. Note that the step ``Find candidate findings $fs_{\mathrm{candidate}}$ using Algorithm \ref{alg:candidate_findings}'' is optional, because checking other findings always yields $U(f')=0$. The main reason for this whole step is to avoid these unnecessary computations. Also note that the \emph{for} loop in Algorithm \ref{alg:bed}, the most computationally intensive step, is parallelizable, which makes the algorithm highly scalable.

\begin{algorithm}
\SetAlgoLined
\caption{Bayesian experimental design algorithm for differential diagnosis using KL divergence}
\label{alg:bed}
 \kwSetup{QMR belief network with all the probabilities, maximum step N, utility threshold $U_{\mathrm{thresh}}$, a case with positive findings $fs$, and disease $d$} 
 \kwInput{An initial set of positive findings $f^+$ and negative findings $f^-$.}
\kwOutput{Diagnosis $d_{\mathrm{diagnosis}}$}
 \kwInit{current step n = 0} 
 \While{n \textless= N}{
  \eIf{n = N}
  {Make final diagnosis $d_{\mathrm{diagnosis}}$ using Algorithm \ref{alg:bi}\;\Return}{
  n = n + 1\;
  Find candidate findings $fs_{\mathrm{candidate}}$ using Algorithm \ref{alg:candidate_findings}\;
  Set $U_{\mathrm{max}}=0$ and $f_{\mathrm{next}}=null$\;
  \For{$f_{\mathrm{candidate}}$ in $fs_{\mathrm{candidate}}$} {
    Compute utility function $U$ for $f_{\mathrm{candidate}}$ using equation \ref{eq:utility-kl}\;
    \If{ $U > U_{\mathrm{max}}$}{ $U_{\mathrm{max}} = U, f_{\mathrm{next}} = f_{\mathrm{candidate}}$\; }
  }
  \eIf{ $U_{\mathrm{max}} > U_{\mathrm{thresh}}$}
  {Make final diagnosis $d_{\mathrm{diagnosis}}$ using Algorithm \ref{alg:bi}\;\Return}
  {Inquire finding $f_{\mathrm{next}}$ and add it to the positive findings $f^+$ or negative findings $f^-$ according to the answer}
  }
 }
\end{algorithm}

\begin{algorithm}
\SetAlgoLined
\caption{Algorithm for identifying candidate findings}
\label{alg:candidate_findings}
 \kwSetup{Prepare a hashmap $\mathcal{M}_{\mathrm{fd}}$ that maps each finding to associated diseases and a hashmap $\mathcal{M}_{\mathrm{df}}$ that maps each disease to associated findings from the QMR belief network} 
 \kwInput{Currently known positive findings $f^+$ and negative findings $f^-$.}
\kwOutput{Candidate findings $fs_{\mathrm{candidate}}$}
\kwInit{Hashtable $\mathcal{F}$ to contain candidate findings} 
 \For{$f$ in $f^+$}{
    Get candidate diseases $d_{\mathrm{candidate}}=\mathcal{M}_{\mathrm{fd}}[f]$\;
    \For{$d$ in $d_{\mathrm{candidate}}$}{
        Add candidate findings $\mathcal{M}_{\mathrm{df}}[d]$ to $\mathcal{F}$ 
    }
 }
Remove findings from $\mathcal{F}$ if they are in $f^+$ or $f^-$\;
\Return $\mathcal{F}$
\end{algorithm}

\subsection{Multi-step Look Ahead Search for Finding Inquiry}
The finding inquiry in the previous section is a one step look ahead search. In this section, we extend it to multi-step look ahead search. In particular, instead of comparing the disease distribution $P(d_j|f^+,f^-)$ and $P(d_j|f^+,f^-,f'{=}y)$ as in equation \ref{eq:utility-kl}, in N-step look ahead search, we compare $P(d_j|f^+,f^-)$ and $P(d_j|f^+,f^-,f'{=}y,f_{\mathrm{opt}}^{(2)}, f_{\mathrm{opt}}^{(3)},\cdots,f_{\mathrm{opt}}^{(N)} )$. It can be implemented in a recursive manner. For the base case, the $N$th step optimal action
\begin{equation*}
    f_{\mathrm{opt}}^{(N)} = \argmax_{f^{(N)}}U(f^{(N)}|f^{(1)},f^{(2)},\cdots,f^{(N-1)})
\end{equation*}
For the other cases, the $n$th step optimal action $f_{\mathrm{opt}}^{(n)}$ can be computed by 
\begin{align*}
&\argmax_{f^{(n)}}U(f^{(n)}|f^{(1)},\cdots,f^{(n-1)},f_{\mathrm{opt}}^{(n+1)},\cdots,f_{\mathrm{opt}}^{(N)}) \\
=&\argmax_{f^{(n)}}\\
&\displaystyle\sum_{y=0,1}P(f^{(n)}{=}y|f^+,f^-,f^{(1)},\cdots,f^{(n-1)})\sum_{j}D_{KL}\big(\\
&P(d_j|f^+,f^-,f^{(1)},\cdots,f^{(n)}{=}y,f_{\mathrm{opt}}^{(n+1)},\cdots,f_{\mathrm{opt}}^{(N)})||\\
&P(d_j|f^+,f^-,f^{(1)},f^{(2)},\cdots,f^{(n-1)})\big)
\end{align*}
\section{Experiments}
\subsection{Datasets}
\subsubsection{SymCAT and HPO}
SymCAT is a medical symptom checker online and mobile application developed by Ahead Research. This dataset summarizes a large amount of data from the Center for Disease Control and Prevention (CDC) is what we refer to as the ``SymCAT dataset'' in this paper. Another dataset we use in our experiment is the Human Phenotype Ontology (HPO) \cite{pmid33264411}, an ontology of phenotypic abnormalities for human disease, containing over 13,000 terms and over 156,000 annotations to hereditary diseases. 

Note that these two datasets naturally have the form of a QMR belief network, except that the marginal probability of each disease is missing. In our experiments, we assume every disease has the same marginal probability, specifically, in the case of $n$ diseases for example, the marginal probabilities of each disease are all equal to $\frac{1}{n}$. Here's a snapshot of the SymCAT dataset: 

\begin{tcolorbox}
``abdominal-aortic-aneurysm'': [
    [``Sharp abdominal pain'', 0.53], 
    [``Back pain'',0.35], 
    [``Shortness of breath'',0.28] 
], \\
``abdominal-hernia'': [ 
    [``Sharp abdominal pain'',0.65], 
    [``Groin mass'',0.32], 
    [``Ache all over'',0.29], 
    [``Upper abdominal pain'',0.23] 
]
\end{tcolorbox}

``abdominal-aortic-aneurysm'' and ``abdominal-hernia'' are two different diseases. ``Back pain'' is a finding for ``abdominal-aortic-aneurysm''. The number ``0.35'' means the disease ``abdominal-aortic-aneurysm'' has a probability of 0.35 to cause the finding ``Back pain''. Note that all the diseases are distinct, but some findings are commonly shared under diseases. In this snapshot, the finding ``Sharp abdominal pain'' is a finding for both ``abdominal-aortic-aneurysm'' and ``abdominal-hernia'', though the corresponding probability is different.

\subsubsection{Muzhi and Dxy}
The Muzhi dataset \cite{wei-etal-2018-task} and the Dxy (also named DX) dataset \cite{xu2019endtoend} are two medical dialogue datasets contain 710 and 527 conversations that are collected from the pediatric department in two Chinese online healthcare communities, where patients chat with doctor online. Table \ref{table:dialogue-stats} shows the statistics of these two datasets. They not only contain dialog data, but also include the normalization of the symptom utterance and summarization of the cases using these normalized symptoms. To demonstrate the summarization of the cases, we show an example of a case after translating Chinese to English and organizing the structure with minimal changes, 

\begin{tcolorbox}
``disease'': ``URTI'', \\
``explicitly informed findings'': [``Cough'':True, ``Running Nose'':True, ``Nasal congestion'':True, ``Sneeze'':True], \\
``implicitly informed findings'': [``Phlegm'':False, ``Fever'':False]
\end{tcolorbox}

In this case, the patient informs the doctor that he/she has symptoms like ``Cough'' ``Running Nose'', ``Nasal congestion'', and ``Sneeze''. Then the doctor asks two questions, ``Do you have Phlegm?'' and ``Do you have Fever?'', and the patient's answer to both questions is negative.

\begin{table}[!ht]
\centering
\renewcommand{\arraystretch}{1.2}
{
\resizebox{0.9\linewidth}{!}{
\small
\begin{tabular}{l|*{3}{r}}
\toprule
Statistics & Dxy & Muzhi \\
\hline
Number of diseases & 5 & 4 \\
Number of symptoms & 41 & 66 \\
Number of cases in training set & 423 & 568 \\
Number of cases in test set  & 104 & 142 \\
\bottomrule
\end{tabular}
}
}
\caption{Statistics of dialogue datasets}
\label{table:dialogue-stats}
\end{table}




\subsubsection{Data characteristics}
Intuitively, besides the discrepancy in training set and test set for raw dialogue datasets, there are some other important characteristics that can represent the difficulty of the data. Here ``difficulty'' is loosely defined as the following: a dataset is more difficult than another when the accuracy is lower than the other when both are solved by the same algorithm. These characteristics include the total number of disease, total number of findings, number of diseases per finding, i.e. the average number of diseases that can cause a finding, and number of findings per disease, i.e. the average number of findings that can be caused by a disease. Table \ref{table:all-stats} summarizes these statistics. Although HPO has the highest number of total diseases and findings, the number of diseases per finding is much smaller than the SymCAT dataset. In fact, due to the low number of diseases per finding, given some observed findings, one can rule out many irrelevant diseases, no matter how many total diseases there are. We argue that the number of diseases per finding is the most dominant factor to measure the difficulty of a dataset, and our experiments also show some evidence. 

\begin{table}[!ht]
\centering
\renewcommand{\arraystretch}{1.2}
{
\resizebox{1.0\linewidth}{!}{
\small
\begin{tabular}{l|*{3}{r}}
\toprule
Dataset(\#Diseases) &\#Findings &\#Diseases per finding  &\#Findings per disease  \\
\hline
SymCAT (200) & 328 & 7.003 & 11.485 \\
SymCAT (300) & 349 & 9.865 & 11.477 \\
SymCAT (400) & 355 & 12.910 & 11.458 \\
HPO (500)  & 1901 & 2.547 & 9.682 \\
HPO (1000) & 3599 & 4.497 & 16.183 \\
Dxy train set (41) & 41 & 3.122 & 25.6 \\
Muzhi train set (66) & 355 & 3.394 & 56.0 \\
\bottomrule
\end{tabular}
}
}
\caption{Statistics of all datasets}
\label{table:all-stats}
\end{table}

\subsection{Experiment Settings}
There are two important hyperparameters in our BED (Bayesian experimental design) algorithm, namely the threshold and maximum steps. We ran a grid search over them by setting the threshold for the utility function to [0.01, 0.05, 0.10] and maximum steps to [10, 15, 20]. We simulate 10,000 cases using the QMR belief network and measure the top 1, top 3, top 5 accuracy, and average steps. Our code is written in Python 3 \cite{python3}. And all experiments are run on a 2.20 GHz Intel Xeon CPU. \footnote{We make our code available at \url{https://github.com/hguan6/differential_diagnosis}}

We adopt the simulation process in \citet{he2020fit}. Each simulated case has one true disease and starts with one informed positive finding. A subset of diseases in the SymCAT and HPO datasets is used to simulate the cases. Specifically, for SymCAT, a subset of 200, 300, 400 diseases are tested; and for HPO, a subset of 500, 1000 diseases are tested. For multi-step look ahead search, we experiment with one-step and two-step search.

For dialogue datasets Muzhi \cite{wei-etal-2018-task} and Dxy \cite{xu2019endtoend}, instead of directly using the training set to train a model, we use it to build a QMR belief network, which is then used by our Bayesian experimental design method. When determining the probability for the edges, we use empirical mean of the number of positive findings for each disease. We use two settings for the marginal probability of each disease, one is according to the data distribution of the training set, the other is to assume even distribution over all diseases, i.e, 25\% for all diseases in Muzhi and 20\% for all diseases in Dxy. Due to the small dataset size and the discrepancy between the training set and test set, we found no significant advantage of either of the settings. For the rest of the paper, we assume the marginal probabilities are equal for all diseases in the same dataset.

\subsection{Results}
Table \ref{table:symcat-result} and Table \ref{table:hpo-result} show some representative results for the SymCAT dataset and HPO dataset respectively. These results are selected from all the combination of the hyperparameters to demonstrate the advantage of our method over prior work REFUEL \cite{refuel} and FIT \cite{he2020fit}, while the full results are shown in the Appendix. For the SymCAT dataset, though FIT achieves slightly higher accuracy than REFUEL, it uses five to six more steps on average. However, our BED (Bayesian Experimental Design) algorithm outperforms FIT by using a similar or less number of steps. For the HPO dataset, our BED method outperforms REFUEL and FIT by a large margin even with a much smaller number of steps. This may be a result of a small number of diseases per finding in the HPO dataset, as is shown in the previous analysis of the data characteristics. The experiment results also show that the two-step search further improves the accuracy of diagnosis using less steps. The accuracy of the two-step search is even close to the accuracy of a “cheater solver”, where we made all the findings observable to the inference engine. The accuracy of the cheater solver for the SymCAT dataset and the HPO dataset are shown in Table \ref{table:cheater}. Since the cheater solver has access to all the information that any inquiry strategy can obtain, it always gives the best ``guess'' to each case and its accuracy can be viewed as the upper bound of accuracy for any data simulated by the QMR belief network.

\begin{table*}[!ht]
\centering
\renewcommand{\arraystretch}{1.2}
{
\resizebox{1.0\linewidth}{!}{
\small
\begin{tabular}{c|cccc|cccc|cccc|cccc}
\toprule
 \multirow{2}{*}{\#Disease} &  \multicolumn{4}{c}{REFUEL} & \multicolumn{4}{|c}{FIT} & 
 \multicolumn{4}{|c}{One-step BED(ours)} &
 \multicolumn{4}{|c}{Two-step BED(ours)}\\
\cmidrule(lr){2-5} \cmidrule(lr){6-9} \cmidrule(lr){10-13} \cmidrule(lr){14-17}
~& Top1 & Top3 & Top5 & Steps & Top1 & Top3 & Top5 & Steps & Top1 & Top3 & Top5 & Steps & Top1 & Top3 & Top5 & Steps\\ 
 \hline
200 & 53.76 & 73.12 & 79.53 & 8.24 & 55.65 & 80.71 & 89.32 & 12.02 & 57.89 & 80.71 & 89.17 & 8.89 & 71.49 & 90.70 & 95.75 & 8.14\\
300 & 47.65 & 66.22 & 71.79 & 8.39 & 48.23 & 73.82 & 84.21 & 13.10 & 50.10 & 73.44 & 83.58 & 8.71 & 67.37 & 86.61 & 92.57 & 6.86\\ 
400 & 43.01 & 59.65 & 68.89 & 8.92 & 44.63 & 69.22 & 69.54 & 14.86 & 46.90 & 70.50 & 80.68 & 10.28 & 64.78 & 83.12 & 89.35 & 7.11\\
\bottomrule
\end{tabular}
}
}
\caption{Top 1, 3, 5 accuracy (\%) and average steps of REFUEL, FIT, and BED for the SymCAT dataset. Higher accuracy is better and lower average steps is better.}
\label{table:symcat-result}
\end{table*}

\begin{table*}[!ht]
\centering
\renewcommand{\arraystretch}{1.2}
{
\resizebox{1.0\linewidth}{!}{
\small
\begin{tabular}{c|cccc|cccc|cccc|cccc}
\toprule
 \multirow{2}{*}{\# Disease} &  \multicolumn{4}{c}{REFUEL} & \multicolumn{4}{|c}{FIT} & 
 \multicolumn{4}{|c}{One-step BED(ours)} &
 \multicolumn{4}{|c}{Two-step BED(ours)} \\
\cmidrule(lr){2-5} \cmidrule(lr){6-9} \cmidrule(lr){10-13} \cmidrule(lr){14-17}
~& Top1 & Top3 & Top5 & Steps & Top1 & Top3 & Top5 & Steps & Top1 & Top3 & Top5 & Steps & Top1 & Top3 & Top5 & Steps\\ 
 \hline
500 & 64.33 & 73.14 & 75.34 & 8.09 & 76.23 & 84.17 & 86.99 & 5.34 & 98.14 & 99.46 & 99.62 & 4.93 & 98.69 & 99.85 & 99.91 & 2.67\\
1000 & 40.08 & 62.67 & 67.42 & 14.19 & 68.65 & 77.23 & 80.35 & 11.09 & 96.49 & 98.21 & 98.69 & 6.72 & 98.25 & 99.60 & 99.75 & 3.16\\
\bottomrule
\end{tabular}
}
}
\caption{Top 1, 3, 5 accuracy (\%) and average steps of REFUEL, FIT, and BED for the HPO dataset. Higher accuracy is better and lower average steps is better.}
\label{table:hpo-result}
\end{table*}

\begin{table}[!ht]
\centering
\renewcommand{\arraystretch}{1.2}
{
\resizebox{0.8\linewidth}{!}{
\small
\begin{tabular}{c|*{4}{c}}
\toprule
Dataset(\#Diseases) & Top1 & Top3 & Top5  \\
\hline
SymCAT (200) & 71.83 & 91.19 & 96.22 \\
SymCAT (300) & 69.31 & 87.87 & 94.06 \\
SymCAT (400) & 66.43 & 85.66 & 91.98 \\
HPO (500) & 98.91 & 99.97 & 100.0 \\
HPO (1000) & 99.34 & 99.97 & 100.0 \\
\bottomrule
\end{tabular}
}
}
\caption{Top 1, 3, 5 accuracy (\%) of Cheater solver for the Muzhi dataset and Dxy dataset}
\label{table:cheater}
\end{table}

Table \ref{table:muzhi-dxy-result} compares the result of the top 1 accuracy using REFUEL \cite{refuel}, GAMP \cite{Xia_Zhou_Shi_Lu_Huang_2020}, FIT \cite{he2020fit} and BED (Our method) for Muzhi and Dxy datasets. These two are very small datasets as is shown in Table \ref{table:dialogue-stats}, whose  main  purpose  is  for  dialogue  generation. GAMP and FIT perform well on these two datasets because their systems just need to mimic the previous actions. We found that, however, there is a significant difference between the training set and test set so that the QMR belief network built from the training set can not predict the data in the test set with high accuracy. In particular, in both of these two datasets, there are three symptoms that each of them is associated with a disease in the test set but is not associated with the same disease in the training set. \citet{he2020fit} also observed discrepancies in these two datasets. We argue that, if one can collect a big real life diagnosis dataset even without the inquiry steps, the estimation of the probability numbers in the QMR belief network becomes much more accurate, and our approach will have very strong performance.

\begin{table}[!ht]
\centering
\renewcommand{\arraystretch}{1.2}
{
\resizebox{0.9\linewidth}{!}{
\small
\begin{tabular}{c|*{5}{c}}
\toprule
Dataset & REFUEL & GAMP & FIT & BED (Ours)  \\
\hline
Muzhi & 71.8 & 73 & 72.6 & 65.5 \\
Dxy & 75.7 & 76.9 & 81.1 & 80.8 \\
\bottomrule
\end{tabular}
}
}
\caption{Top 1 accuracy (\%) of REFUEL, GAMP, FIT, and BED for the Muzhi dataset and Dxy dataset}
\label{table:muzhi-dxy-result}
\end{table}
\section{Discussions}
The dialogue datasets Dxy and Muzhi are very small, which do not take advantage of the real power of our method because the QMR belief networks built from such small datasets do not reflect reality. In clinical practice, healthcare providers can define their own set of findings and diseases, and collect findings/disease information from patient records to build their own QMR belief network without violation of privacy. In that case, the QMR belief network is customized to the real clinical practice. When large data is not available, the QMR belief network can be initialized by human experts according to their domain knowledge and experience, and be updated over time as more data is collected.

Our approach has some practical advantages in production. Firstly, in previous work based on reinforcement learning and partial variational autoencoders, one has to re-train their model when new findings or new diseases are added or the probability numbers in the network change. Our approach does not require any training, which in turns makes it adapt to changes in the QMR belief network without additional effort. Secondly, our approach can make inference in any step and allows the user to manually set the findings to any state in any step. When a doctor uses our system as an assistant, he/she does not have to follow the inquiry strategy all the way to the end. Instead, they can manually change the question to ask and determine when to stop the algorithm. Lastly, our method allows clinicians to reason about every decision made by our method by inspecting the graphical structure of the QMR belief network. 

In this work, we simplified the inference step by assuming diseases are mutually exclusive, i.e. one disease per case. To generalize to multiple diseases, one can simply use the general case of Bayesian inference during the inference phase.
Another arguable limitation of our approach is that our inquiry phase does not directly optimize the ultimate objective, that is, maximizing the prediction accuracy. Ideally, one can achieve it by replacing Bayesian experimental design by reinforcement learning in the inquiry phase. However, one should be aware of the cost and difficulty of training a reinforcement agent and adaptation to new changes of the QMR belief network. 
Note that we compute the exact probabilities of each disease in our algorithm, which can be expensive when the number of diseases is large. Thus our future work is to enable faster inference and inquiry without exact probability computations.

\section{Conclusions}
In this work, we combine reasoning with the QMR belief network and Bayesian experimental design to come up with a simple, scalable, and effective approach for automated differential diagnosis of diseases.
Experiments verify the superiority of our approach, in which we achieve SOTA accuracy with less inquiry steps on two simulated datasets and competitive results on two small dialogue datasets. Moreover, our multi-step look ahead search method further improves the inquiry phase, achieving much higher accuracy with less inquiry steps, which is even on par with the inference accuracy with fully observed information.

\bibliography{refs}

\begin{thebibliography}{26}
\providecommand{\natexlab}[1]{#1}

\bibitem[{Chaloner and Verdinelli(1995)}]{10.1214/ss/1177009939}
Chaloner, K.; and Verdinelli, I. 1995.
\newblock {Bayesian Experimental Design: A Review}.
\newblock \emph{Statistical Science}, 10(3): 273 -- 304.

\bibitem[{Cimino and Shortliffe(2006)}]{biomedical_informatics}
Cimino, J.~J.; and Shortliffe, E.~H. 2006.
\newblock \emph{Biomedical Informatics: Computer Applications in Health Care
  and Biomedicine (Health Informatics)}.
\newblock Berlin, Heidelberg: Springer-Verlag.
\newblock ISBN 0387289860.

\bibitem[{He et~al.(2021)He, Mao, Ma, Huang, Hernández-Lobato, and
  Chen}]{he2020fit}
He, W.; Mao, X.; Ma, C.; Huang, Y.; Hernández-Lobato, J.~M.; and Chen, T.
  2021.
\newblock FIT: a Fast and Accurate Framework for Solving Medical Inquiring and
  Diagnosing Tasks.
\newblock arXiv:2012.01065.

\bibitem[{Jaakkola and Jordan(1999)}]{Jaakkola_1999}
Jaakkola, T.~S.; and Jordan, M.~I. 1999.
\newblock Variational Probabilistic Inference and the QMR-DT Network.
\newblock \emph{Journal of Artificial Intelligence Research}, 10: 291–322.

\bibitem[{Kao, Tang, and Chang(2018)}]{kao2018context}
Kao, H.-C.; Tang, K.-F.; and Chang, E. 2018.
\newblock Context-aware symptom checking for disease diagnosis using
  hierarchical reinforcement learning.
\newblock In \emph{Proceedings of the AAAI Conference on Artificial
  Intelligence}, volume~32.

\bibitem[{Kononenko(2001)}]{kononenko2001machine}
Kononenko, I. 2001.
\newblock Machine learning for medical diagnosis: history, state of the art and
  perspective.
\newblock \emph{Artificial Intelligence in medicine}, 23(1): 89--109.

\bibitem[{Köhler et~al.(2021)Köhler, Gargano, Matentzoglu, Carmody,
  Lewis-Smith, Vasilevsky, Danis, Balagura, Baynam, Brower, Callahan, Chute,
  Est, Galer, Ganesan, Griese, Haimel, Pazmandi, Hanauer, Harris, Hartnett,
  Hastreiter, Hauck, He, Jeske, Kearney, Kindle, Klein, Knoflach, Krause,
  Lagorce, McMurry, Miller, Munoz-Torres, Peters, Rapp, Rath, Rind, Rosenberg,
  Segal, Seidel, Smedley, Talmy, Thomas, Wiafe, Xian, Yüksel, Helbig, Mungall,
  Haendel, and Robinson}]{pmid33264411}
Köhler, S.; Gargano, M.; Matentzoglu, N.; Carmody, L.~C.; Lewis-Smith, D.;
  Vasilevsky, N.~A.; Danis, D.; Balagura, G.; Baynam, G.; Brower, A.~M.;
  Callahan, T.~J.; Chute, C.~G.; Est, J.~L.; Galer, P.~D.; Ganesan, S.; Griese,
  M.; Haimel, M.; Pazmandi, J.; Hanauer, M.; Harris, N.~L.; Hartnett, M.~J.;
  Hastreiter, M.; Hauck, F.; He, Y.; Jeske, T.; Kearney, H.; Kindle, G.; Klein,
  C.; Knoflach, K.; Krause, R.; Lagorce, D.; McMurry, J.~A.; Miller, J.~A.;
  Munoz-Torres, M.~C.; Peters, R.~L.; Rapp, C.~K.; Rath, A.~M.; Rind, S.~A.;
  Rosenberg, A.~Z.; Segal, M.~M.; Seidel, M.~G.; Smedley, D.; Talmy, T.;
  Thomas, Y.; Wiafe, S.~A.; Xian, J.; Yüksel, Z.; Helbig, I.; Mungall, C.~J.;
  Haendel, M.~A.; and Robinson, P.~N. 2021.
\newblock {{T}he {H}uman {P}henotype {O}ntology in 2021}.
\newblock \emph{Nucleic Acids Res}, 49(D1): D1207--D1217.

\bibitem[{Ledley and Lusted(1959)}]{ledley1959reasoning}
Ledley, R.~S.; and Lusted, L.~B. 1959.
\newblock Reasoning foundations of medical diagnosis.
\newblock \emph{Science}, 130(3366): 9--21.

\bibitem[{Ma et~al.(2019)Ma, Tschiatschek, Palla, Lobato, Nowozin, and
  Zhang}]{ma2019eddi}
Ma, C.; Tschiatschek, S.; Palla, K.; Lobato, J. M.~H.; Nowozin, S.; and Zhang,
  C. 2019.
\newblock {EDDI}: Efficient Dynamic Discovery of High-Value Information with
  Partial {VAE}.

\bibitem[{Miller, Masarie, and Myers(1986)}]{pmid3537611}
Miller, R.; Masarie, F.~E.; and Myers, J.~D. 1986.
\newblock {{M}{D} {C}omput{Q}uick medical reference ({Q}{M}{R}) for diagnostic
  assistance}.
\newblock \emph{MD Comput}, 3(5): 34--48.

\bibitem[{Miller(1994)}]{10.1136/jamia.1994.95236141}
Miller, R.~A. 1994.
\newblock {Medical Diagnostic Decision Support Systems—Past, Present, And
  Future: A Threaded Bibliography and Brief Commentary}.
\newblock \emph{Journal of the American Medical Informatics Association}, 1(1):
  8--27.

\bibitem[{Miller et~al.(1986)Miller, McNeil, Challinor, Masarie, and
  Myers}]{pmid3544509}
Miller, R.~A.; McNeil, M.~A.; Challinor, S.~M.; Masarie, F.~E.; and Myers,
  J.~D. 1986.
\newblock {{W}est {J} {M}ed{T}he {I}{N}{T}{E}{R}{N}{I}{S}{T}-1/{Q}{U}{I}{C}{K}
  {M}{E}{D}{I}{C}{A}{L} {R}{E}{F}{E}{R}{E}{N}{C}{E} project--status report}.
\newblock \emph{West J Med}, 145(6): 816--822.

\bibitem[{Ng and Jordan(1999)}]{ng1999approximate}
Ng, A.~Y.; and Jordan, M.~I. 1999.
\newblock Approximate Inference A lgorithms for Two-Layer Bayesian Networks.
\newblock In \emph{NIPS}, 533--539. Citeseer.

\bibitem[{Pearl(1988)}]{pearl1988}
Pearl, J. 1988.
\newblock \emph{Probabilistic Reasoning in Intelligent Systems: Networks of
  Plausible Inference}.
\newblock San Francisco, CA, USA: Morgan Kaufmann Publishers Inc.
\newblock ISBN 1558604790.

\bibitem[{Peng et~al.(2018)Peng, Tang, Lin, and Chang}]{refuel}
Peng, Y.-S.; Tang, K.-F.; Lin, H.-T.; and Chang, E. 2018.
\newblock REFUEL: Exploring Sparse Features in Deep Reinforcement Learning for
  Fast Disease Diagnosis.
\newblock In Bengio, S.; Wallach, H.; Larochelle, H.; Grauman, K.;
  Cesa-Bianchi, N.; and Garnett, R., eds., \emph{Advances in Neural Information
  Processing Systems}, volume~31. Curran Associates, Inc.

\bibitem[{Ranschaert, Morozov, and Algra(2019)}]{ai_imaging}
Ranschaert, E.~R.; Morozov, S.; and Algra, P.~R. 2019.
\newblock \emph{Artificial Intelligence in Medical Imaging: Opportunities,
  Applications and Risks}.
\newblock Springer Publishing Company, Incorporated, 1st edition.
\newblock ISBN 3319948776.

\bibitem[{Shwe and Cooper(1991)}]{pmid1743005}
Shwe, M.; and Cooper, G. 1991.
\newblock {{C}omput {B}iomed {R}es{A}n empirical analysis of
  likelihood-weighting simulation on a large, multiply connected medical belief
  network}.
\newblock \emph{Comput Biomed Res}, 24(5): 453--475.

\bibitem[{Shwe et~al.(1991)Shwe, Middleton, Heckerman, Henrion, Horvitz,
  Lehmann, and Cooper}]{pmid1762578}
Shwe, M.~A.; Middleton, B.; Heckerman, D.~E.; Henrion, M.; Horvitz, E.~J.;
  Lehmann, H.~P.; and Cooper, G.~F. 1991.
\newblock {{M}ethods {I}nf {M}ed{P}robabilistic diagnosis using a reformulation
  of the {I}{N}{T}{E}{R}{N}{I}{S}{T}-1/{Q}{M}{R} knowledge base. {I}. {T}he
  probabilistic model and inference algorithms}.
\newblock \emph{Methods Inf Med}, 30(4): 241--255.

\bibitem[{Sutton et~al.(2020)Sutton, Pincock, Baumgart, Sadowski, Fedorak, and
  Kroeker}]{sutton2020overview}
Sutton, R.~T.; Pincock, D.; Baumgart, D.~C.; Sadowski, D.~C.; Fedorak, R.~N.;
  and Kroeker, K.~I. 2020.
\newblock An overview of clinical decision support systems: benefits, risks,
  and strategies for success.
\newblock \emph{NPJ digital medicine}, 3(1): 1--10.

\bibitem[{Tang et~al.(2016)Tang, Kao, Chou, and Chang}]{tang2016inquire}
Tang, K.-F.; Kao, H.-C.; Chou, C.-N.; and Chang, E.~Y. 2016.
\newblock Inquire and diagnose: Neural symptom checking ensemble using deep
  reinforcement learning.
\newblock In \emph{NIPS Workshop on Deep Reinforcement Learning}.

\bibitem[{Van~Rossum and Drake(2009)}]{python3}
Van~Rossum, G.; and Drake, F.~L. 2009.
\newblock \emph{Python 3 Reference Manual}.
\newblock Scotts Valley, CA: CreateSpace.
\newblock ISBN 1441412697.

\bibitem[{Wei et~al.(2018)Wei, Liu, Peng, Tou, Chen, Huang, Wong, and
  Dai}]{wei-etal-2018-task}
Wei, Z.; Liu, Q.; Peng, B.; Tou, H.; Chen, T.; Huang, X.; Wong, K.-f.; and Dai,
  X. 2018.
\newblock Task-oriented Dialogue System for Automatic Diagnosis.
\newblock In \emph{Proceedings of the 56th Annual Meeting of the Association
  for Computational Linguistics (Volume 2: Short Papers)}, 201--207. Melbourne,
  Australia: Association for Computational Linguistics.

\bibitem[{Xia et~al.(2020)Xia, Zhou, Shi, Lu, and
  Huang}]{Xia_Zhou_Shi_Lu_Huang_2020}
Xia, Y.; Zhou, J.; Shi, Z.; Lu, C.; and Huang, H. 2020.
\newblock Generative Adversarial Regularized Mutual Information Policy Gradient
  Framework for Automatic Diagnosis.
\newblock \emph{Proceedings of the AAAI Conference on Artificial Intelligence},
  34(01): 1062--1069.

\bibitem[{Xu et~al.(2019)Xu, Zhou, Gong, Liang, Tang, and Lin}]{xu2019endtoend}
Xu, L.; Zhou, Q.; Gong, K.; Liang, X.; Tang, J.; and Lin, L. 2019.
\newblock End-to-End Knowledge-Routed Relational Dialogue System for Automatic
  Diagnosis.
\newblock \emph{Proceedings of the AAAI Conference on Artificial Intelligence},
  33: 7346--7353.

\bibitem[{Yu et~al.(2003)Yu, Tu, Tu, and Pattipati}]{feili2003}
Yu, F.; Tu, F.; Tu, H.; and Pattipati, K. 2003.
\newblock Multiple disease (fault) diagnosis with applications to the QMR-DT
  problem.
\newblock In \emph{SMC'03 Conference Proceedings. 2003 IEEE International
  Conference on Systems, Man and Cybernetics. Conference Theme - System
  Security and Assurance (Cat. No.03CH37483)}, volume~2, 1187--1192 vol.2.

\bibitem[{Yu et~al.(2007)Yu, Tu, Tu, and Pattipati}]{yu_2007}
Yu, F.; Tu, F.; Tu, H.; and Pattipati, K.~R. 2007.
\newblock A Lagrangian Relaxation Algorithm for Finding the MAP Configuration
  in QMR-DT.
\newblock \emph{IEEE Transactions on Systems, Man, and Cybernetics - Part A:
  Systems and Humans}, 37(5): 746--757.

\end{thebibliography}

\appendix
\section{More experiment results}

We show the full results for the SymCAT dataset using one-step look ahead search in Table \ref{table:symcat200}, \ref{table:symcat300}, \ref{table:symcat400} and the full results using two-step look ahead search in Table \ref{table:symcat200_two-step}, \ref{table:symcat300_two-step}, \ref{table:symcat400_two-step}. Table \ref{table:hpo500}, \ref{table:hpo1000} are the full results for the HPO dataset using one-step look ahead search and Table \ref{table:hpo500_two-step}, \ref{table:hpo1000_two-step} are the full results using two-step look ahead search. The results in the main content of our paper are selected from these tables. We compare our results with REFUEL \cite{refuel} and FIT \cite{he2020fit}.

\begin{table*}[!ht]
\centering
\renewcommand{\arraystretch}{1.2}
{
\resizebox{1.0\linewidth}{!}{
\small
\begin{tabular}{c|cccc|cccc|cccc}
\toprule
 \multirow{2}{*}{Max steps} &  \multicolumn{4}{c}{Threshold=0.01} & \multicolumn{4}{|c}{Threshold=0.05} & 
 \multicolumn{4}{|c}{Threshold=0.1} \\
\cmidrule(lr){2-5} \cmidrule(lr){6-9} \cmidrule(lr){10-13}
~& Top1 & Top3 & Top5 & Steps & Top1 & Top3 & Top5 & Steps & Top1 & Top3 & Top5 & Steps\\ 
 \hline
10 & 57.89 & 80.71 & 89.17 & 8.89 & 56.89 & 80.53 & 89.02 & 8.35 & 54.78 & 79.36 & 88.39 & 7.31\\
15 & 63.91 & 85.66 & 92.31 & 12.35 & 62.76 & 85.21 & 92.20 & 10.69 & 58.51 & 82.42 & 90.62 & 8.48\\ 
20 & 67.62 & 87.80 & 93.76 & 14.83 & 64.58 & 86.57 & 93.20 & 12.07 & 58.97 & 82.80 & 91.38 & 9.08\\
\bottomrule
\end{tabular}
}
}
\caption{Top 1, 3, 5 accuracy and average steps for SymCAT (200) using one-step look ahead search. Higher accuracy is better and lower average steps is better.}
\label{table:symcat200}
\end{table*}

\begin{table*}[!ht]
\centering
\renewcommand{\arraystretch}{1.2}
{
\resizebox{1.0\linewidth}{!}{
\small
\begin{tabular}{c|cccc|cccc|cccc}
\toprule
 \multirow{2}{*}{Max steps} &  \multicolumn{4}{c}{Threshold=0.01} & \multicolumn{4}{|c}{Threshold=0.05} & 
 \multicolumn{4}{|c}{Threshold=0.1} \\
\cmidrule(lr){2-5} \cmidrule(lr){6-9} \cmidrule(lr){10-13}
~& Top1 & Top3 & Top5 & Steps & Top1 & Top3 & Top5 & Steps & Top1 & Top3 & Top5 & Steps\\ 
 \hline
10 & 70.60 & 89.49 & 95.01 & 7.07 & 70.42 & 89.59 & 94.70 & 6.77 & 70.93 & 90.20 & 95.28 & 6.39\\
15 & 71.25 & 90.21 & 95.33 & 9.79 & 72.03 & 90.76 & 95.79 & 9.02 & 71.49 & 90.70 & 95.75 & 8.14\\ 
20 & 71.21 & 90.67 & 95.75 & 11.87 & 72.27 & 91.29 & 96.34 & 10.39 & 70.76 & 90.17 & 95.77 & 9.08\\
\bottomrule
\end{tabular}
}
}
\caption{Top 1, 3, 5 accuracy and average steps for SymCAT (200) using two-step look ahead search. Higher accuracy is better and lower average steps is better.}
\label{table:symcat200_two-step}
\end{table*}

\begin{table*}[!ht]
\centering
\renewcommand{\arraystretch}{1.2}
{
\resizebox{1.0\linewidth}{!}{
\small
\begin{tabular}{c|cccc|cccc|cccc}
\toprule
 \multirow{2}{*}{Max steps} &  \multicolumn{4}{c}{Threshold=0.01} & \multicolumn{4}{|c}{Threshold=0.05} & 
 \multicolumn{4}{|c}{Threshold=0.1} \\
\cmidrule(lr){2-5} \cmidrule(lr){6-9} \cmidrule(lr){10-13}
~& Top1 & Top3 & Top5 & Steps & Top1 & Top3 & Top5 & Steps & Top1 & Top3 & Top5 & Steps\\ 
 \hline
10 & 51.59 & 73.89 & 83.58 & 9.16 & 50.10 & 73.44 & 83.58 & 8.71 & 47.25 & 70.89 & 81.36 & 7.83\\
15 & 57.22 & 78.78 & 86.13 & 12.93 & 56.11 & 78.46 & 87.25 & 11.52 & 52.45 & 75.66 & 85.61 & 9.48\\ 
20 & 62.54 & 82.02 & 90.37 & 15.94 & 59.16 & 81.50 & 89.77 & 13.31 & 53.36 & 76.36 & 86.13 & 10.10\\
\bottomrule
\end{tabular}
}
}
\caption{Top 1, 3, 5 accuracy and average steps for SymCAT (300) using one-step look ahead. Higher accuracy is better and lower average steps is better.}
\label{table:symcat300}
\end{table*}

\begin{table*}[!ht]
\centering
\renewcommand{\arraystretch}{1.2}
{
\resizebox{1.0\linewidth}{!}{
\small
\begin{tabular}{c|cccc|cccc|cccc}
\toprule
 \multirow{2}{*}{Max steps} &  \multicolumn{4}{c}{Threshold=0.01} & \multicolumn{4}{|c}{Threshold=0.05} & 
 \multicolumn{4}{|c}{Threshold=0.1} \\
\cmidrule(lr){2-5} \cmidrule(lr){6-9} \cmidrule(lr){10-13}
~& Top1 & Top3 & Top5 & Steps & Top1 & Top3 & Top5 & Steps & Top1 & Top3 & Top5 & Steps\\ 
 \hline
10 & 66.9 & 86.15 & 92.50 & 7.37 & 66.82 & 86.17 & 92.81 & 7.13 & 67.37 & 86.61 & 92.57 & 6.86\\
15 & 69.47 & 87.80 & 93.92 & 10.26 & 68.01 & 86.8 & 93.04 & 9.67 & 67.08 & 86.69 & 92.90 & 8.75\\ 
20 & 68.07 & 87.18 & 93.24 & 12.78 & 67.68 & 86.91 & 93.47 & 11.38 & 68.06 & 86.99 & 93.27 & 9.82\\
\bottomrule
\end{tabular}
}
}
\caption{Top 1, 3, 5 accuracy and average steps for SymCAT (300) using two-step look ahead. Higher accuracy is better and lower average steps is better.}
\label{table:symcat300_two-step}
\end{table*}

\begin{table*}[!ht]
\centering
\renewcommand{\arraystretch}{1.2}
{
\resizebox{1.0\linewidth}{!}{
\small
\begin{tabular}{c|cccc|cccc|cccc}
\toprule
 \multirow{2}{*}{Max steps} &  \multicolumn{4}{c}{Threshold=0.01} & \multicolumn{4}{|c}{Threshold=0.05} & 
 \multicolumn{4}{|c}{Threshold=0.1} \\
\cmidrule(lr){2-5} \cmidrule(lr){6-9} \cmidrule(lr){10-13}
~& Top1 & Top3 & Top5 & Steps & Top1 & Top3 & Top5 & Steps & Top1 & Top3 & Top5 & Steps\\ 
 \hline
10 & 44.25 & 66.57 & 77.11 & 9.39 & 44.55 & 67.53 & 77.47 & 9.05 & 42.39 & 65.21 & 75.85 & 8.22\\
15 & 52.54 & 73.95 & 82.96 & 13.27 & 51.26 & 73.60 & 82.95 & 12.15 & 46.90 & 70.50 & 80.68 & 10.26\\ 
20 & 57.34 & 78.25 & 86.68 & 16.67 & 55.94 & 76.94 & 85.56 & 14.29 & 49.40 & 73.45 & 82.93 & 11.05\\
\bottomrule
\end{tabular}
}
}
\caption{Top 1, 3, 5 accuracy and average steps for SymCAT (400) using one-step look ahead. Higher accuracy is better and lower average steps is better.}
\label{table:symcat400}
\end{table*}

\begin{table*}[!ht]
\centering
\renewcommand{\arraystretch}{1.2}
{
\resizebox{1.0\linewidth}{!}{
\small
\begin{tabular}{c|cccc|cccc|cccc}
\toprule
 \multirow{2}{*}{Max steps} &  \multicolumn{4}{c}{Threshold=0.01} & \multicolumn{4}{|c}{Threshold=0.05} & 
 \multicolumn{4}{|c}{Threshold=0.1} \\
\cmidrule(lr){2-5} \cmidrule(lr){6-9} \cmidrule(lr){10-13}
~& Top1 & Top3 & Top5 & Steps & Top1 & Top3 & Top5 & Steps & Top1 & Top3 & Top5 & Steps\\ 
 \hline
10 & 64.78 & 83.12 & 89.35 & 7.11 & 64.73 & 83.01 & 89.44 & 7.35 & 64.99 & 83.44 & 89.83 & 7.51\\
15 & 65.62 & 84.69 & 90.98 & 10.57 & 65.83 & 84.38 & 90.82 & 9.94 & 65.61 & 84.67 & 90.84 & 9.27\\ 
20 & 65.26 & 84.36 & 91.10 & 13.03 & 65.97 & 84.56 & 91.58 & 11.93 & 65.35 & 84.53 & 91.04 & 10.41\\
\bottomrule
\end{tabular}
}
}
\caption{Top 1, 3, 5 accuracy and average steps for SymCAT (400) using two-step look ahead. Higher accuracy is better and lower average steps is better.}
\label{table:symcat400_two-step}
\end{table*}

\begin{table*}[!ht]
\centering
\renewcommand{\arraystretch}{1.2}
{
\resizebox{1.0\linewidth}{!}{
\small
\begin{tabular}{c|cccc|cccc|cccc}
\toprule
 \multirow{2}{*}{Max steps} &  \multicolumn{4}{c}{Threshold=0.01} & \multicolumn{4}{|c}{Threshold=0.05} & 
 \multicolumn{4}{|c}{Threshold=0.1} \\
\cmidrule(lr){2-5} \cmidrule(lr){6-9} \cmidrule(lr){10-13}
~& Top1 & Top3 & Top5 & Steps & Top1 & Top3 & Top5 & Steps & Top1 & Top3 & Top5 & Steps\\ 
 \hline
10 & 95.64 & 97.99 & 98.66 & 4.30 & 95.19 & 97.78 & 98.49 & 4.13 & 94.77 & 97.76 & 98.38 & 3.95\\
15 & 97.46 & 99.05 & 99.33 & 4.69 & 97.23 & 99.05 & 99.36 & 4.46 & 96.58 & 98.83 & 99.27 & 4.22\\ 
20 & 98.14 & 99.46 & 99.62 & 4.93 & 97.98 & 99.36 & 99.59 & 4.52 & 97.48 & 99.56 & 99.74 & 4.25\\
\bottomrule
\end{tabular}
}
}
\caption{Top 1, 3, 5 accuracy and average steps for HPO (500) using one-step look ahead. Higher accuracy is better and lower average steps is better.}
\label{table:hpo500}
\end{table*}

\begin{table*}[!ht]
\centering
\renewcommand{\arraystretch}{1.2}
{
\resizebox{1.0\linewidth}{!}{
\small
\begin{tabular}{c|cccc|cccc|cccc}
\toprule
 \multirow{2}{*}{Max steps} &  \multicolumn{4}{c}{Threshold=0.01} & \multicolumn{4}{|c}{Threshold=0.05} & 
 \multicolumn{4}{|c}{Threshold=0.1} \\
\cmidrule(lr){2-5} \cmidrule(lr){6-9} \cmidrule(lr){10-13}
~& Top1 & Top3 & Top5 & Steps & Top1 & Top3 & Top5 & Steps & Top1 & Top3 & Top5 & Steps\\ 
 \hline
10 & 98.69 & 99.85 & 99.91 & 2.67 & 98.42 & 99.72 & 99.87 & 2.66 & 98.35 & 99.84 & 99.94 & 2.56\\
15 & 98.87 & 99.81 & 99.88 & 2.97 & 98.87 & 99.86 & 99.90 & 2.81 & 98.68 & 99.88 & 99.96 & 2.72\\ 
20 & 98.84 & 99.92 & 99.95 & 3.03 & 99.05 & 99.92 & 99.97 & 2.90 & 98.62 & 99.83 & 99.94 & 2.77\\
\bottomrule
\end{tabular}
}
}
\caption{Top 1, 3, 5 accuracy and average steps for HPO (500) using two-step look ahead. Higher accuracy is better and lower average steps is better.}
\label{table:hpo500_two-step}
\end{table*}

\begin{table*}[!ht]
\centering
\renewcommand{\arraystretch}{1.2}
{
\resizebox{1.0\linewidth}{!}{
\small
\begin{tabular}{c|cccc|cccc|cccc}
\toprule
 \multirow{2}{*}{Max steps} &  \multicolumn{4}{c}{Threshold=0.01} & \multicolumn{4}{|c}{Threshold=0.05} & 
 \multicolumn{4}{|c}{Threshold=0.1} \\
\cmidrule(lr){2-5} \cmidrule(lr){6-9} \cmidrule(lr){10-13}
~& Top1 & Top3 & Top5 & Steps & Top1 & Top3 & Top5 & Steps & Top1 & Top3 & Top5 & Steps\\ 
 \hline
10 & 89.97 & 93.65 & 95.22 & 5.96 & 89.90 & 94.29 & 95.79 & 5.70 & 89.77 & 93.95 & 95.51 & 5.49\\
15 & 94.30 & 96.56 & 97.32 & 6.80 & 94.42 & 97.10 & 97.80 & 6.44 & 93.88 & 96.61 & 97.48 & 6.16\\ 
20 & 96.24 & 98.08 & 98.58 & 7.31 & 96.49 & 98.21 & 98.69 & 6.72 & 95.78 & 98.06 & 98.55 & 6.47\\
\bottomrule
\end{tabular}
}
}
\caption{Top 1, 3, 5 accuracy and average steps for HPO (1000) using one-step look ahead. Higher accuracy is better and lower average steps is better.}
\label{table:hpo1000}
\end{table*}

\begin{table*}[!ht]
\centering
\renewcommand{\arraystretch}{1.2}
{
\resizebox{1.0\linewidth}{!}{
\small
\begin{tabular}{c|cccc|cccc|cccc}
\toprule
 \multirow{2}{*}{Max steps} &  \multicolumn{4}{c}{Threshold=0.01} & \multicolumn{4}{|c}{Threshold=0.05} & 
 \multicolumn{4}{|c}{Threshold=0.1} \\
\cmidrule(lr){2-5} \cmidrule(lr){6-9} \cmidrule(lr){10-13}
~& Top1 & Top3 & Top5 & Steps & Top1 & Top3 & Top5 & Steps & Top1 & Top3 & Top5 & Steps\\ 
 \hline
10 & 98.90 & 99.75 & 99.80 & 3.21 & 98.25 & 99.60 & 99.75 & 3.16 & 98.50 & 99.75 & 99.80 & 3.17\\
15 & 98.75 & 99.88 & 99.93 & 3.56 & 98.68 & 99.83 & 99.90 & 3.43 & 98.95 & 99.85 & 99.88 & 3.36\\ 
20 & 99.15 & 99.88 & 99.93 & 3.70 & 98.98 & 99.9 & 99.93 & 3.50 & 98.75 & 99.78 & 99.88 & 3.49\\
\bottomrule
\end{tabular}
}
}
\caption{Top 1, 3, 5 accuracy and average steps for HPO (1000) using two-step look ahead. Higher accuracy is better and lower average steps is better.}
\label{table:hpo1000_two-step}
\end{table*}
\end{document}